\documentclass[10pt,twocolumn,letterpaper]{article}

\usepackage{cvpr}
\usepackage{times}
\usepackage{epsfig}
\usepackage{graphicx}
\usepackage{amsmath}
\usepackage{amssymb}
\usepackage{svg}
\usepackage{color}
\usepackage{array,multirow}
\usepackage{subfig}
\usepackage[linesnumbered,ruled]{algorithm2e}
\usepackage{comment}
\usepackage[pagebackref=true,breaklinks=true,letterpaper=true,colorlinks,bookmarks=false]{hyperref}

\cvprfinalcopy % *** Uncomment this line for the final submission

\ifcvprfinal\fi
\begin{document}

%%%%%%%%% TITLE
\title{Multi-task Learning of Hierarchical Vision-Language Representation}
% Densely Symmetric Attention

\author{Duy-Kien Nguyen$^{1}$~ and ~~Takayuki Okatani$^{1,2}$\\
% \vspace*{-3mm}
$^1$Tohoku University ~~~~ $^2$RIKEN Center for AIP\\
% 6-3-09 Aoba, Aramaki-aza Aoba-ku, Sendai, 980-8579, Japan \\
{\tt\small \{kien, okatani\}@vision.is.tohoku.ac.jp}}

\maketitle
\thispagestyle{empty}

%%%%%%%%% ABSTRACT
\begin{abstract} 
It is still challenging to build an AI system that can perform tasks that involve vision and language at human level. So far, researchers have singled out individual tasks separately, for each of which they have designed networks and trained them on its dedicated datasets. Although this approach has seen a certain degree of success, it comes with difficulties of understanding relations among different tasks and transferring the knowledge learned for a task to others. We propose a multi-task learning approach that enables to learn vision-language representation that is shared by many tasks from their diverse datasets. The representation is hierarchical, and prediction for each task is computed from the representation at its corresponding level of the hierarchy. We show through experiments that our method consistently outperforms previous single-task-learning methods on image caption retrieval, visual question answering, and visual grounding. We also analyze the learned hierarchical representation by visualizing attention maps generated in our network.
\end{abstract}

%%%%%%%%% BODY TEXT
\section{Introduction}
Since the recent successes of deep learning on single modality tasks, multi-modal tasks lying on the intersection of vision and language, such as image captioning \cite{mscoco, Young_2014_TACL}, visual question answering (VQA) \cite{balanced_vqa_v2, VQA}, visual grounding \cite{Plummer_2017_IJCV} etc., have attracted increasing attention in the related fields. Despite the fact that each of these tasks has basically been studied independently of others, there must be close connections among them. For instance, each task may occupy a different level in a hierarchy of sub-tasks comprising the cognitive function associated with vision and language. To gain deeper understanding of such hidden relations among these vision-language tasks, we think that multi-task learning of these tasks will be a promising direction of research.

Although we have seen significant progresses in multi-task learning of unimodal tasks of vision \cite{Kendall_2018_CVPR, Ray_2018_ECCV} or language \cite{Luong_2016_ICLR, Uddin_2018_ICLR, Subramanian_2018_ICLR} so far, there has been only a limited amount of progress in multi-task learning of vision-language tasks. This may be attributable to the diversity of these tasks. In addition to differences in inputs and outputs of the tasks, 
their level of complexity differs, too. Even though these tasks share some structures in common, it is unclear how to learn them in the framework of multi-task learning. 

A solution to this difficulty is to create and use a dataset designed for multi-task learning, where multiple objectives are given to identical inputs. Indeed, recent studies follow this approach, in which they have gained early successes by joint training of different vision-language tasks using multiple objectives, such as answer and question generation for VQA \cite{Li_2018_CVPR}, and caption and scene graph generation for image captioning \cite{Li_2017_CVPR}. 
However, this approach cannot be employed when such dedicated datasets are not available. It may be impossible to create such datasets for arbitrary combinations of vision-language tasks. To do this, we need to limit the range of tasks, which makes it hard for the learned results to generalize to other tasks or datasets.

In this paper, aiming to resolve these issues, we propose a framework for joint learning of 
multiple vision-language tasks. Our goal is to enable to learn vision-language representation that is shared by many tasks from their diverse data sources. To make this possible, we employ \textit{Dense Co-attention} layers, which were developed for VQA and shown to perform competitively with existing methods \cite{Nguyen_2018_CVPR}. Using a stack of Dense Co-attention layers, we can gradually update the visual and linguistic features at each layer, in which their fine-grained interaction is considered at the level of individual image regions and words. We utilize this property to learn hierarchical vision-language representation such that each individual task takes the learned representation at a different level of the hierarchy corresponding to its complexity. We design a network consisting of an encoder for computing shared hierarchical representation and multiple task-specific decoders for making prediction from the representation; see Fig.~\ref{fig:diagram}. This design enables multi-task learning from diverse data sources; to be rigorous, we train the same network alternately on each task/dataset based on a scheduling algorithm.  

We evaluate this method on three vision-language tasks, image caption retrieval, visual question answering, and visual grounding, using popular datasets, Flickr30K captions \cite{Young_2014_TACL}, MS-COCO captions \cite{mscoco}, VQA 2.0 \cite{balanced_vqa_v2}, and Flickr30K-Entities \cite{Plummer_2017_IJCV}. The results show that our method outperforms previous ones that are trained on individual tasks and datasets. We also visualize the internal behaviours of the task-specific decoders to analyze 
effects of joint learning of the multiple tasks. 

%------------------------------------------------------------------------
\section{Related Work}

\paragraph{Vision-language representation learning}
Recently, studies of multi-modal tasks of vision and language have made significant progress, such as image captioning \cite{Anderson_2018_CVPR, Lu_2018_CVPR}, visual question answering \cite{Nguyen_2018_CVPR, Teney_2018_CVPR}, visual grounding \cite{Yeh_2017_NIPS, Plummer_2017_IJCV}, image caption retrieval \cite{Hyeonseob_2017_CVPR, Huang_2018_CVPR}, and visual dialog \cite{visdial}. In the last few years, researchers have demonstrated the effectiveness of learning representations shared by the two modalities in a supervised fashion. However, these studies deal with a single task at a time.

\paragraph{Transfer learning} 
A basic method of transfer learning in deep learning is to train a neural network for a source task and use it in some ways for a target task. This method has been successful in a wide range of problems in computer vision and natural language processing. For multi-modal vision-language problems, early works explored similar approaches that used pretrained models trained on some source tasks. Plummer et al. \cite{Plummer_2017_IJCV} proposed to use a pretrained network trained on a visual grounding task to enrich the shared representational space of images and captions, improving accuracy of image caption retrieval. Lin et al. \cite{Lin_2016_ECCV} proposed to use pretrained models of VQA and CQA (caption Q\&A); they compute answer predictions for multiple questions and then treat them as features of an image and a caption, computing relevance between them. In this study, instead of transferring knowledge from a source task to a target task in a single direction (e.g., via pretrained models), we consider a multi-task learning framework in which learning multiple tasks will be mutually beneficial to each individual task. This is made possible by the proposed network and its training methodology; it suffices only to train our network for individual tasks with their loss functions and supervised data. 

\paragraph{Multi-task learning of vision-language tasks} Since its introduction \cite{Caruana1997}, multi-task learning has achieved many successes in several areas including computer vision and natural language processing. However, there have been only a few works that explored joint learning of multiple multi-modal tasks of vision and language. Li et al. \cite{Li_2017_CVPR} proposed a method for learning relations between multiple regions in the image by jointly refining the features of three different semantic tasks, scene graph generation, object detection, and image/region captioning. Li et al. \cite{Li_2018_CVPR} showed that joint training on VQA and VQG (visual question generation) contributes to improve VQA accuracy and also understanding of interactions among images, questions, and answers. Although these works have demonstrated the potential of multi-task learning for the vision-language tasks, they strongly rely on the availability of the datasets providing supervision over multiple tasks, where an input is shared by all the tasks while a different label is given to it for each task.

%------------------------------------------------------------------------
\section{Learning Vision-Language Interaction}

\subsection{Problem Formulation}

We consider multiple vision-language tasks, in each of which an output $O$ is to be estimated from an input pair of $I$ and $S$, where $I$ is an image and $S$ is a sentence. The input pair $I$ and $S$ have the same formats for all the tasks (with differences in the interpretation of $S$ for different tasks), whereas the output $O$ will naturally be different for each task. For example, in VQA, $O$ is a set of 
confident scores of answers to the input question $S$ in a predefined answer set; in image caption retrieval, $O$ is a set of binary values indicating the relevance of the input caption $S$; in visual grounding, $O$ is a set of binary variables specifying a set of image regions corresponding to the phrases in the input sentence $S$.

The input image is represented by a set of region features, which we denote by $I = [i_1, ..., i_T]$; in our experiments, we use a bag of region features from a pretrained Faster-RCNN \cite{Teney_2018_CVPR}. The input sentence is represented by a sequence $S = [s_1, ..., s_N]$ of word features, which are obtained by first computing GloVe embedding vectors \cite{Pennington_2014_EMNLP} of the input words and then inputting them to a two-layer bidirectional LSTM.

\begin{figure}[t]
\centering
\includegraphics[width=80mm]{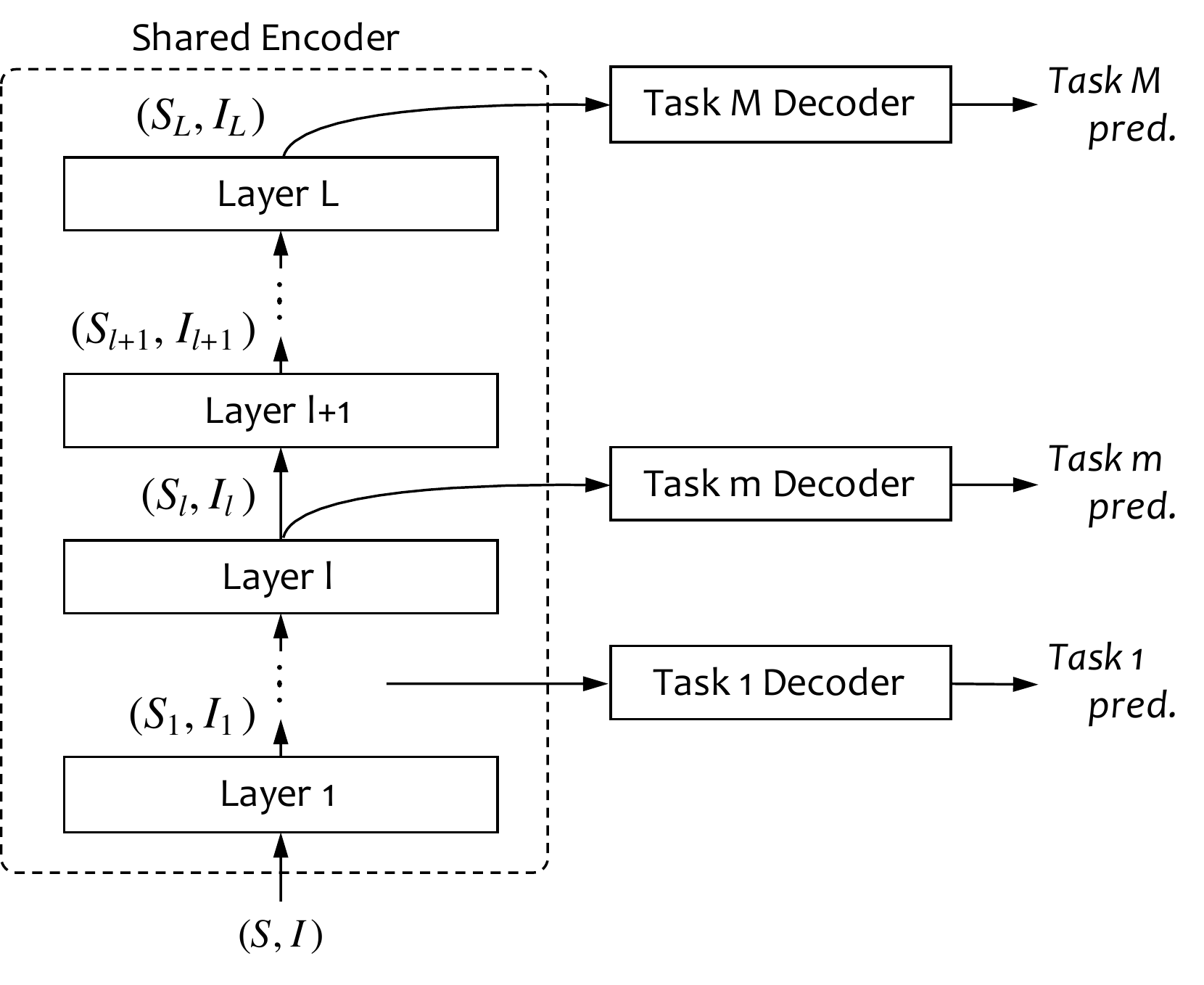}
\caption{The proposed network consists of a shared encoder and task-specific decoders. The shared encoder is a stack of $L$ Dense Co-attention layers and computes hierarchical representation of the input sentence and image. Each of $M$ task-specific decoders receives one intermediate-layer representation to compute prediction for its task. } 
\label{fig:diagram}
\end{figure}

An overview of the proposed network architecture is shown in Fig.~\ref{fig:diagram}. 
It consists of a single encoder shared by all the tasks and multiple task-specific decoders. We will describe these two components below.

\subsection{Shared Encoder}

To construct the shared encoder, we employ the {\em Dense Co-attention} layer \cite{Nguyen_2018_CVPR}. We conjecture that different tasks require different levels of vision-language fusion. Thus, we stack multiple Dense Co-attention layers to extract hierarchical, fused features of the input image $I$ and sentence $S$. We attach a decoder for each task to the layer that is the best fit for the task in terms of the fusion hierarchy, as shown in Fig.~\ref{fig:diagram}. 

Starting with $S_0=S$ and $I_0=I$, the shared encoder incrementally updates the language and vision features at each Dense Co-attention layer as
\begin{equation}
(S_{l}, I_{l}) = \text{DCL}_l(S_{l-1}, I_{l-1}),
\end{equation}
where 
$S_{l} = [s_{l,1}, ..., s_{l,N}] \in \mathbb{R}^{d\times N}$, 
$I_{l} = [i_{l,1}, ..., i_{l,T}] \in \mathbb{R}^{d\times T}$, $S_{l-1} = [s_{l-1,1}, ..., s_{l-1,N}] \in \mathbb{R}^{d\times N}$, and $I_{l-1} = [i_{l-1,1}, ..., i_{l-1,T}] \in \mathbb{R}^{d\times T}$; $\text{DCL}_l$ indicates the input-output function realized by the $l$-th Dense Co-attention layer.
In each Dense Co-attention layer, two attention maps are generated in a symmetric fashion, i.e., the one over image regions conditioned on each sentence word and the other over sentence words conditioned on each image region, where multiplicative attention is employed. The generated attention maps are then applied to $I_{l-1}$ and $S_{l-1}$ to yield  $\hat{I}_{l-1}$ and $\hat{S}_{l-1}$, respectively. Finally, the original and attended features of image and sentence are fused by first concatenating them and then applying a linear transform followed by ReLU. This is done for sentence feature and image feature, respectively, as
\begin{align}
s_{l,n} &= \text{ReLU}\left(W_{S_l}\begin{bmatrix}s_{l-1,n}\\ \hat{i}_{l-1,n}\end{bmatrix} + b_{S_l}\right) + s_{l-1,n},\\
i_{l,t} &= \text{ReLU}\left(W_{I_l}\begin{bmatrix}i_{l-1,t}\\ \hat{s}_{l-1,t}\end{bmatrix} + b_{I_l}\right) + i_{l-1,t},
\end{align}
where $W_{S_l}\in \mathbb{R}^{d \times 2d}$, $b_{S_l} \in\mathbb{R}^d$, $W_{I_l} \in \mathbb{R}^{d\times 2d}$ and $b_{I_l}\in\mathbb{R}^d$ are learnable parameters.

\subsection{Task-specific Decoders}

As shown in Fig.~\ref{fig:diagram}, we design a task-specific decoder for each task and attach it to the layer of the shared encoder selected for the task. Letting $l$ be the index of this layer, the decoder receives $(S_l, I_l)$ and produces the final output $O$ for this task. We explain below its design for each of the three tasks considered in this study.

\subsubsection{Image Caption Retrieval}

In this task, we calculate the relevant score for the input pair ($I$, $S$). The decoder for this task consists of two summary networks and a scoring layer. Let $l_\mathrm{R}$ be the index of the layer of the shared encoder to which this decoder is attached. The first summary network computes a vector  $v_{I_{l_\mathrm{R}}}\in \mathbb{R}^d$ that summarizes the image features $I_{l_\mathrm{R}}=[i_{l_\mathrm{R},1},\ldots,i_{l_\mathrm{R},T}]$ of $T$ regions. The second summary network computes $v_{S_{l_\mathrm{R}}}\in \mathbb{R}^d$ that summarizes the sentence features $S_{l_\mathrm{R}}=[s_{l_\mathrm{R},1},\ldots,s_{l_\mathrm{R,}N}]$ of $N$ words.

The two summary networks have the same architecture. Let us take the image summary network for explanation. It consists of a two-layer feedforward network that yields attention maps over the $T$ image regions and a mechanism that applies the attention maps to $I_{l_\mathrm{R}}$ to obtain the summary vector $v_{I_{l_\mathrm{R}}}$. 

The feedforward network has $d$ hidden units with ReLU non-linearity, which receives the image feature of a single region and outputs $K$ scores. To be specific, denoting the feedforward network by $\text{MLP}_I$, it maps the feature vector of each region $t(=1,\ldots,T)$ to $K$ scores as
\begin{equation}
c^I_t = [c^I_{1,t}, \ldots, c^I_{K,t}] = \text{MLP}_I(i_{l_\mathrm{R},t}),\;\;t=1,\ldots, T.
\end{equation}

These scores are then normalized by softmax across the $T$ regions to obtain $K$ parallel attention maps over the $T$ regions, which are  averaged to produce the final attention map $[\alpha^I_1,\ldots,\alpha^I_T]$; more specifically,
\begin{equation}
\alpha^I_t = \frac{1}{K} \sum_{k=1}^K \frac{\exp(c^I_{k,t})}{\sum_{t=1}^{T}\exp(c^I_{k,t})},\;\;t=1,\ldots, T.
\end{equation}
The summary vector $v_{I_{l_\mathrm{R}}}$ is the weighted sum of the image feature vectors using this attention weights, i.e., 
\begin{equation}
v_{I_{l_\mathrm{R}}} = \sum_{t=1}^{T} \alpha^I_t i_{l_\mathrm{R},t},
\end{equation}
As mentioned above, we generate $K$ parallel attention maps and average them to obtain a single attention map. This is to capture more diverse attention distribution. 

We follow the same procedure to compute the summary vector $v_{S_{l_\mathrm{R}}}$, where a two-layer feedforward network $\mathrm{MLP}_S$ generating $K$ parallel attention maps over $N$ word features $S_{l_\mathrm{R}} = [s_{l_\mathrm{R},1}, \ldots, s_{l_\mathrm{R},N}]$ is used. Using the two summary vectors $v_{I_{l_\mathrm{R}}}$ and $v_{S_{l_\mathrm{R}}} \in \mathbb{R}^{d}$ thus obtained, 
the scoring layer computes the relevant score of an image-caption pair $(I, S)$ as
\begin{equation}
\mbox{score} (I, S) = \sigma(v_{I_{l_\mathrm{R}}}^{\top} W v_{S_{l_\mathrm{R}}}),
\end{equation}
where $\sigma$ is the logistic function and $W \in \mathbb{R}^{d \times d}$ is a learnable weight matrix.

\subsubsection{Visual Question Answering}

In this task, we compute the scores of a set of predefined answers for the input image-question pair. Let $l_{\mathrm{Q}}$ be the index of the layer to which the decoder is attached. We employ the same architectural design as in the decoder for image caption retrieval to compute the summary vectors $v_{I_{l_\mathrm{Q}}}$ and $v_{S_{l_\mathrm{Q}}}$ from the input features, i.e.,  $S_{l_\mathrm{Q}}=[s_{l_{\mathrm{Q}},1},\ldots,s_{l_{\mathrm{Q}},N}] $ and $I_{l_\mathrm{Q}}=[i_{l_{\mathrm{Q}},1},\ldots,i_{l_{\mathrm{Q}},T} ]$. To obtain these summary vectors, two summary networks, each of which is two-layer feedforward network with $d$ hidden units and ReLU nonlinearity, are used to compute $K$ attention maps, and then they  are applied to the input features.

Following \cite{Nguyen_2018_CVPR}, we compute scores for a set of the predefined answers by using a two-layer feedforward network having $d$ hidden units with ReLU non-linearity and  output units for the scores; the output units employ the logistic function for their activation function. Denoting the network by $\mathrm{MLP}$, the scores are calculated as
\begin{equation}
(\mbox{scores of answers}) = \sigma \Big(\text{MLP}\big(\begin{bmatrix}v_{I_{l_\mathrm{Q}}}\\ v_{S_{l_\mathrm{Q}}}\end{bmatrix}\big)\Big).
\end{equation}

\subsubsection{Visual Grounding}
\label{subsec:visual_grounding}

This is a task in which given an image and a phrase (usually one contained in a caption describing the image), we want to identify the image region corresponding to the phrase. Previously proposed approaches attempt to learn to score each region-phrase pair separately; any context in the caption is not taken into account, or any joint inference about global interaction between all phrases in the caption is not performed. We believe that context is important for understanding a local phrase in a sentence, needless to mention its necessity for higher-level tasks.

Let $l_{\mathrm{G}}$ be the index of the layer of the shared encoder connecting to the decoder for this task. Given $P = [(b_{1}, e_{1}), ..., (b_{H}, e_{H})]$ where $(b_{h}, e_{h})$ indicates the start and end indexes of the $h$-th phrase in the input $N$ word caption ($1\leq b_h \leq e_h \leq N$), %$S_{l(vg)}$,
we compute the feature $p_h \in \mathbb{R}^d$ for the $h$-th phrase by pooling the word features in the index range of $[b_h:e_h]$ as
\begin{equation}
p_h = \text{AvgPooling}(S_{l_\mathrm{G}}[b_h:e_h]).
\end{equation}
Here we use average pooling to produce a fixed-size vector representation of a phrase $p_h$. This can also be seen as computing an attended feature using an attention map with equal weights on words in the phrase and zero weights on other words.

We then compute the score for a pair of a phrase $p_h \in \mathbb{R}^d$ and an image region $i_{l_{\mathrm{G}},t} \in \mathbb{R}^d$ as
\begin{equation}
\text{score}(p_h, i_t) = \sigma \Big(p_h^{\top} W i_{l_{\mathrm{G}},t} \Big),
\end{equation}
where $\sigma$ is the logistic function and $W \in \mathbb{R}^{d \times d}$ is a learnable weight matrix.

\section{Training on Multiple Tasks}

We train the proposed network on the above multiple tasks. Considering their diversity, we use a strategy to train it on a single selected task at a time and iterate this by switching between the tasks. (Note that we cannot simultaneously train the network on these tasks by minimizing the sum of their losses, because the inputs differ among tasks.) 

\subsection{Task-switching schedule}
\label{sec:taskschedule}

It is essential for which task and how many times we update the parameters of the network. In this study, we employ two strategies. One is a curriculum learning approach that starts from a single task and increases the number of tasks one by one, i.e., training first on single tasks, then on pairs of tasks, and finally on all tasks. The other is a scheduling method when training more than one task in this curriculum. To be specific, we employ the strategy of periodical task switching as in \cite{Dong_2015_IJCNLP} but with different iterations of parameter updates for each task. Following \cite{Luong_2016_ICLR}, we update the network parameters for $i$-th task for $C\alpha_i$ iterations before switching to a new task, where $C$ is the number of iterations in an updating cycle that we specify; $\alpha_i$ is determined as explained below. Algorithm \ref{alg:multitask} shows the entire procedure. More details are given in the supplementary material.

\subsection{Choosing Layers Best Fit for Tasks}
\label{sec:layer_choice}

We need to decide which layer $l(i)$ of the shared encoder is the best fit for each task $i$. We pose it as a hyperparameter search, in which we also determine other parameters for training each task $i$, i.e., $\text{\# step}_i$ (step size for learning rate decay), the number of iteration $\text{\# iter}_i$ (used to determine $\alpha_i$), and the batch size $\text{bs}_i$. To choose them, we conduct a grid search by training the network on each individual task. 

After that, these hyperparameters are used in joint learning of the tasks. Denoting the number of tasks to be learned by $M'$($=1,\, 2$ or $3$), the step size of training is given by $\text{\# step} = \sum_{i=1}^{M'} \text{\# step}_i$;  the total number of iterations is $\text{\# iter} =\sum_{i=1}^{M'} \text{\# iter}_i $; and $\alpha_i$ is determined as $\alpha_i = \text{\# iter}_i / \text{\# iter}$. The batch size $\text{bs}_i$ and layer $l(i)$ determined as above are fixed in all the subsequent training processes.

\begin{algorithm}[t]
\SetAlgoLined
$\text{num\_cycle} = \lfloor \frac{\text{\# iter}}{C} \rfloor$

S = $\big( [1] * C\alpha_1 + ... + [M'] * C\alpha_{M'} \big) * \text{num\_cycle}$

\# Array operation in Python style: 

\# [1] * 3 + [2] * 2 = [1, 1, 1, 2, 2] 

\For{task i in S}{
1: Sample pairs of an input and output: $\mathbf{x, y} \sim \mathbb{P}_i$

2: $\mathbf{h}_{i\mathbf{x}} \leftarrow %\mathbf{E}_{i\theta}(\mathbf{x})$
\mathbf{E}_{l(i)}(\mathbf{x})$

3: Output prediction $\mathbf{\hat{y}} \leftarrow %\mathbf{D}_{i\theta}(\mathbf{h}_{i\mathbf{x}})$
\mathbf{D}_{i}(\mathbf{h}_{i\mathbf{x}})$

4: $\theta \leftarrow \text{Adam}(\nabla_\theta L(\mathbf{y, \hat{y}}))$
}
\caption{Training the proposed network on $M'$ tasks. 
$\mathbf{E}_l$ 
represents a sub-network of the shared encoder up to $l$-th layer ($l=1,\ldots,L$);   $\mathbf{D}_1,\ldots,\mathbf{D}_{M'}$ are $M'$ task-specific decoders; $\theta$ indicates their parameters. $l(i)$ is   the index of the layer to which the decoder for $i$-th task is attached. We represent the output of this layer for an input $\mathbf{x}$ as $\mathbf{E}_{l(i)}$. 
}
\label{alg:multitask}
\end{algorithm}

\section{Experiments}

We conducted a series of experiments to test the effectiveness of the proposed approach. 

\subsection{Datasets and Evaluation Methods}
\label{sec:data}

\paragraph{Image Caption Retrieval}
We use two datasets for this task, MS-COCO and Flickr30k. MS-COCO consists of 82,783 \emph{train} and 40,504 \emph{val} images. Following the standard procedure  \cite{Karpathy_2015_CVPR}, we use the 1,000 \emph{val} images and the 1,000 or 5,000 \emph{test} images, which are selected from the original 40,504 \emph{val} images. We use all of the 82,783 \emph{train} images for training. Flickr30k consists of 31,783 images collected from Flickr. Following the standard procedure \cite{Karpathy_2015_CVPR}, we split them into \emph{train}, \emph{val}, and \emph{test} sets; \emph{val} and \emph{test} contains 1,000 images for each and \emph{train} contains all the others. We report $\text{Recall}@K (K=1, 5, 10)$ (i.e., recall rates at the top 1, 5, and 10 results).

\paragraph{Visual Question Answering} 
We use VQA 2.0 \cite{balanced_vqa_v2}, which is the most popular and the largest (as of now) dataset for this task. It contains questions and answers for images of MS-COCO. There are 443,757 \emph{train}, 214,354 \emph{val}, and 447,793 \emph{test} questions, respectively. The \emph{train} questions are for \emph{train} images of MS-COCO and \emph{val} and \emph{test} questions are for \emph{val} and \emph{test} images of MS-COCO respectively. Following the standard approach \cite{Teney_2018_CVPR}, we choose correct answers appearing more than 8 times to form the predefined answer pool. We use the accuracy metric presented in the original paper \cite{VQA} in all the experiments.

\paragraph{Visual Grounding} 
For visual grounding task, we evaluate our approach on Flickr30k Entities \cite{Plummer_2017_IJCV}, which contains 244,035 annotations to the image-caption pairs (31,783 images and 158,915 captions) of Flickr30k. It provides correspondence between phrases in a sentence and boxes in an image that represent the same entities. The \textit{train}, \textit{val} and \textit{test} are splitted as in the ICR task. We use 1,000 images for \textit{val} and \textit{test} splits each and the rest for \textit{train} split following \cite{Plummer_2017_IJCV}. The task is to localize the corresponding box(es) to each of the given phrases in a sentence. As proposed in \cite{Plummer_2017_IJCV}, we consider a predicted region to be a correct match with a phrase if it has $\text{IOU} \ge 0.5$ with the ground truth bounding box for that phrase. By treating the phrase as the query to retrieve the regions from the input image, we report $\text{Recall}@K (K=1, 5, 10)$ similar to image caption retrieval (the percentage of queries for which a correct match has rank of at most $K$). 

\paragraph{Avoiding Contamination of Training Samples}

As we train the network by alternately switching the tasks, we need to make sure that there is no contamination between training and testing sets for all the tasks. To make a fair comparison with previous studies of VQA, we need to train the network using both \emph{train} and \emph{val} questions of VQA 2.0, as was done in the previous studies. However, if we use \emph{val} questions in our joint learning framework, our network (i.e., the shared encoder) can see the \emph{val} set of MS-COCO, resulting in contamination of training samples. To avoid this, we use the following procedure: i) we first train the network using all the \emph{train} sets for the three tasks and test it on the \emph{test} sets for ICR and VG; ii) we then train the network (from scratch) using the \emph{train} sets for ICR and VG and \emph{train}+\emph{val} sets for VQA and test it on the \emph{test} sets for VQA. This procedure was employed in the experiments of Sec.~\ref{sec:results}, but not employed in the experiments of Sec.~\ref{sec:combi}, because evaluation was done only on \emph{val} sets for all the tasks.

\subsection{Optimal Layers and Training Parameters}

As explained in Sec.~\ref{sec:layer_choice}, we first train our network on each individual task to find the layers fit for each task along with other training parameters. The results are: $l_\mathrm{R}=3$ (image caption retrieval), $l_\mathrm{Q}=5$ (VQA), and $l_\mathrm{G}=2$ (visual grounding). The training parameters were determined accordingly; see the supplementary material for details. We freeze all these parameters throughout all the experiments.

We note here the training method used in all the experiments. We used the Adam optimizer with the parameters $\alpha = 0.001$, $\beta_1 = 0.9$, $\beta_2 = 0.99$, and $\alpha \text{ decay} = 0.5$. We employed a simple training schedule; we halve the learning rate by ``$\alpha \text{ decay}$'' after each ``\# step'' or step size, which are determined above. All the weights in our network were initialized by the method of Glorot et al. \cite{Glorot_2010}. Dropout is applied with probability of 0.3 and 0.1 over FC layers and LSTM, respectively. The dimension $d$ of the feature space is set to 1024.

\subsection{Effects of Joint Learning of Multiple Tasks}
\label{sec:combi}

\begin{table}
\begin{center}
\caption{Performances for different combinations of the three tasks, VQA, VG(visual grounding), and ICR(image caption retrieval). Accuracy (Acc) is reported for VQA, and Recall@1 (R@1) is reported for VG and ICR; two numbers of ICR are image annotation (upper) and image retrieval (lower), respectively. MS-COCO dataset is used for ICR.}
\label{table:ablation} \footnotesize
\begin{tabular}{|c|c|c|c|}
\hline
Task & VQA (Acc) & ICR (R@1) & VG (R@1) \\
\hline
\hline
\multirow{2}{*}{VQA} & \multirow{2}{*}{65.50} & \multirow{2}{*}{-} & \multirow{2}{*}{-} \\
 & & & \\
\hline
\multirow{2}{*}{ICR} & \multirow{2}{*}{-} & 69.05 & \multirow{2}{*}{-} \\
 & & 56.47 & \\
\hline
\multirow{2}{*}{VG} & \multirow{2}{*}{-} & \multirow{2}{*}{-} & \multirow{2}{*}{58.09} \\
 & & & \\
\hline
\multirow{2}{*}{VQA + ICR} & \multirow{2}{*}{66.24} & 69.52  & \multirow{2}{*}{-} \\
 & & 56.74 & \\
\hline
\multirow{2}{*}{VQA + VG} & \multirow{2}{*}{65.85} & \multirow{2}{*}{-} & \multirow{2}{*}{58.07} \\
 & & & \\
\hline
\multirow{2}{*}{ICR + VG} & \multirow{2}{*}{-} & 69.23 & \multirow{2}{*}{58.28} \\
 & & 57.40 & \\
\hline
VQA + ICR& \multirow{2}{*}{66.35} & 70.43 & \multirow{2}{*}{58.26} \\
  + VG & & 57.50 & \\
\hline
\end{tabular}
\end{center}
\end{table}

\begin{table}
\begin{center}
\caption{Effects of joint training when using different image caption retrieval datasets, MS-COCO and Flickr30k. {\em Single} means that each task is learned individually. }
\label{table:compare} \footnotesize
\begin{tabular}{|c|c|cc|cc|}
\hline
\multirow{2}{*}{Task} & Single & \multicolumn{2}{c|}{MS-COCO} & \multicolumn{2}{c|}{Flickr30k} \\
     & $\!\!$(w/o ICR)$\!\!$ & Single & $\!\!\!$+VQA+VG$\!\!\!$ & Single & $\!\!\!$+VQA+VG$\!\!\!$ \\
\hline
\hline
$\!\!$VQA (Acc)$\!\!$ & 65.50 & - & 66.35 & - & 66.09 \\
\hline
$\!\!$\multirow{2}{*}{ICR (R@1)}$\!\!$ & - &69.05 & 70.43 & 67.16 & 72.07 \\
 & - & 56.47 & 57.50 & 53.17 & 56.42 \\
\hline
$\!\!\!$VG (R@1)$\!\!\!$ & 58.09 & - & 58.26 & - & 58.03 \\
\hline
\end{tabular}
\end{center}
\end{table}

\begin{table*}
\begin{center}
\caption{Results of image annotation and retrieval on the Flickr30K and MSCOCO (1000 testing) datasets.}
\label{table:res_icr} \footnotesize
\begin{tabular}{|c|ccc|ccc|ccc|ccc|}
\hline
\multirow{3}{*}{Method} & \multicolumn{6}{c|}{Flickr30k dataset} & \multicolumn{6}{c|}{MSCOCO dataset} \\\cline{2-13}
 & \multicolumn{3}{c|}{Image Annotation} & \multicolumn{3}{c|}{Image Retrieval} & \multicolumn{3}{c|}{Image Annotation} & \multicolumn{3}{c|}{Image Retrieval} \\\cline{2-13}
 & R@1 & R@5 & R@10 & R@1 & R@5 & R@10 & R@1 & R@5 & R@10 & R@1 & R@5 & R@10  \\
\hline
\hline

RNN+FV \cite{Lev_2016_ECCV} & 34.7 & 62.7 & 72.6 & 26.2 & 55.1 & 69.2 & 40.8 & 71.9 & 83.2 & 29.6 & 64.8 & 80.5 \\

OEM \cite{Ivan_2016_ICLR} & - & - & - & - & - & - & 46.7 & 78.6 & 88.9 & 37.9 & 73.7 & 85.9 \\

VQA \cite{Lin_2016_ECCV} & 33.9 & 62.5 & 74.5 & 24.9 & 52.6 & 64.8 & 50.5 & 80.1 & 89.7 & 37.0 & 70.9 & 82.9 \\

RTP \cite{Plummer_2017_IJCV} & 37.4 & 63.1 & 74.3 & 26.0 & 56.0 & 69.3 & - & - & - & - & - & - \\

DSPE \cite{Wang_2016_CVPR} & 40.3 & 68.9 & 79.9 & 29.7 & 60.1 & 72.1 & 50.1 & 79.7 & 89.2 & 39.6 & 75.2 & 86.9 \\

sm-LSTM \cite{Huang_2017_CVPR} & 42.5 & 71.9 & 81.5 & 30.2 & 60.4 & 72.3 & 53.2 & 83.1 & 91.5 & 40.7 & 75.8 & 87.4 \\

RRF \cite{Liu_2017_ICCV} & 47.6 & 77.4 & 87.1 & 35.4 & 68.3 & 79.9 & 56.4 & 85.3 & 91.5 & 43.9 & 78.1 & 88.6 \\

2WayNet \cite{Aviv_2017_CVPR} & 49.8 & 67.5 & - & 36.0 & 55.6 & - & 55.8 & 75.2 & - & 39.7 & 63.3 & - \\

DAN \cite{Hyeonseob_2017_CVPR} & 55.0 & 81.8 & 89.0 & 39.4 & 69.2 & 79.1 & - & - & - & - & - & - \\

VSE++ \cite{Faghri_2017_arxiv} & 52.9 & 79.1 & 87.2 & 39.6 & 69.6 & 79.5 & 64.6 & 89.1 & 95.7 & 52.0 & 83.1 & 92.0 \\

S-E Model \cite{Huang_2018_CVPR} & 55.5 & 82.0 & 89.3 & 41.1 & 70.5 & 80.1 & 69.9 & \bf{92.9} & \bf{97.5} & 56.7 & 87.5 & 94.8 \\

Ours & \bf{71.6} & \bf{84.6} & \bf{90.8} & \bf{56.1} & \bf{82.9} & \bf{89.4} & \bf{70.2} & 89.2 & 95.9 & \bf{57.4} & \bf{88.4} & \bf{95.6} \\
\hline
\end{tabular}
\end{center}
\end{table*}

\begin{table*}
\begin{center}
\caption{Results of the proposed method along with published results of others on VQA 2.0 with single model.}
\label{table:res_vqa2.0} \footnotesize
\begin{tabular}{|c|c|cccc|cccc|}
\hline
\multirow{2}{*}{Method} & \multirow{2}{*}{Feature} & \multicolumn{4}{c|}{Test-dev} & \multicolumn{4}{c|}{Test-standard} \\\cline{3-10}
 & & Overall & Other & Number & Yes/No & Overall & Other & Number & Yes/No\\ 
\hline
\hline

MCB \cite{Fukui_2016_EMNLP} reported in \cite{balanced_vqa_v2} & \multirow{6}{*}{Resnet} & - & - & - & - & 62.27 & 53.36 & 38.28 & 78.82 \\

MF-SIG-T3 \cite{Chen_2017_ICCV} & & 64.73 & 55.55 & 42.99 & 81.29 & - & - & - & - \\

Adelaide-Teney-MSR \cite{Teney_2018_CVPR} & & 62.07 & 52.62 & 39.46 & 79.20 & 62.27 & 52.59 & 39.77 & 79.32 \\

DCN \cite{Nguyen_2018_CVPR} & & 66.72 & 56.77 & 46.65 & 83.70 & 67.04 & 56.95 & 47.19 & 83.85 \\

Memory-augmented Net \cite{Ma_2018_CVPR} & & - & - & - & - & 62.10 &  52.60 & 39.50 & 79.20 \\

VKMN \cite{Su_2018_CVPR} & & - & - & - & - & 64.36 & 57.79 & 37.90 & 83.70 \\

\hline

Adelaide-Teney-MSR \cite{Teney_2018_CVPR} & \multirow{5}{*}{Faster} & 65.32 & 56.05 & 44.21 & 81.82 & 65.67 & 56.26 & 43.90 & 82.20 \\

DCN \cite{Nguyen_2018_CVPR} in our experiments & \multirow{5}{*}{RCNN} & 68.60 & 58.76 & 50.85 & 84.83 & 68.94 & 58.78 & 51.23 & 85.27 \\

Counting Module \cite{Zhang_2018_ICLR} & & 68.09 & 58.97 & \bf{51.62} & 83.14 & 68.41 & 59.11 & 51.39 & 83.56 \\

MLB + DA-NTN \cite{Bai_2018_ECCV} & & 67.56 & 57.92 & 47.14 & 84.29 & 67.94 & 58.20 & 47.13 & 84.60 \\

Ours & & \bf{69.28} & \bf{59.17} & 51.54 & \bf{85.80} & \bf{69.57} & \bf{59.27} & \bf{51.46} & \bf{86.17} \\
\hline
\end{tabular}
\end{center}
\end{table*}

\begin{figure*}
\centering

\begin{minipage}{1\textwidth}
\centering

\begin{minipage}{0.49\textwidth}
\centering
\begin{minipage}{0.47\textwidth}
\centering
\includegraphics[width=\linewidth]{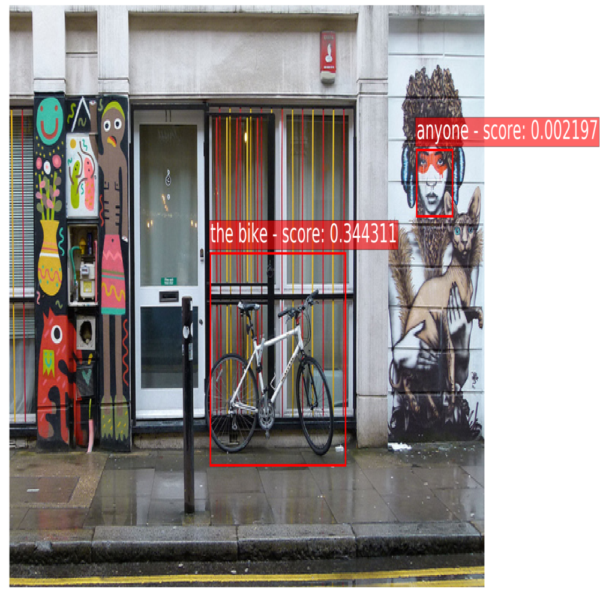}

{\footnotesize is [anyone] riding [the bike]\par}
\end{minipage}
\begin{minipage}{0.47\textwidth}
\centering
\includegraphics[width=\linewidth]{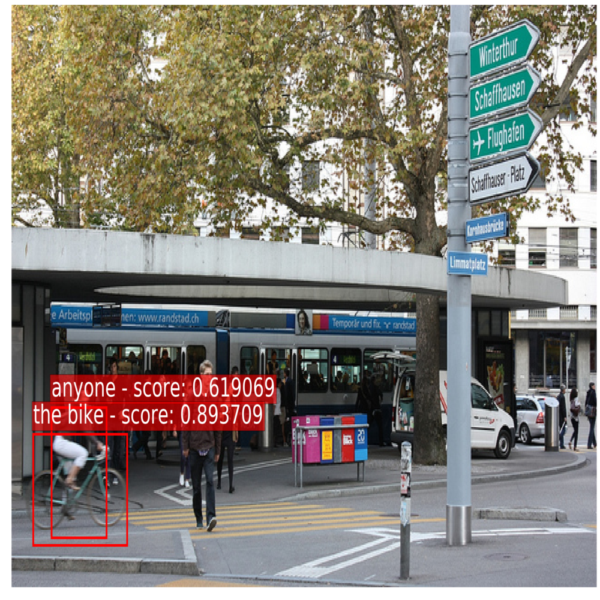}

{\footnotesize is [anyone] riding [the bike]\par}
\end{minipage}
\end{minipage}
\begin{minipage}{0.49\textwidth}
\centering
\begin{minipage}{0.47\textwidth}
\centering
\includegraphics[width=\linewidth]{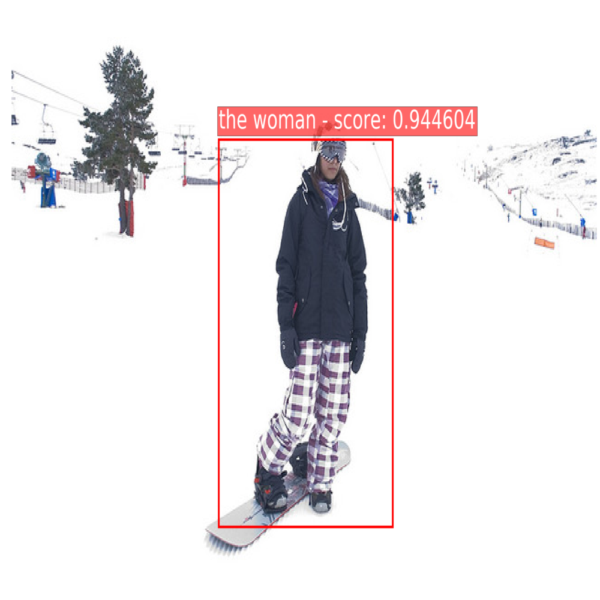}

{\footnotesize what is [the woman] riding\par}
\end{minipage}
\begin{minipage}{0.47\textwidth}
\centering
\includegraphics[width=\linewidth]{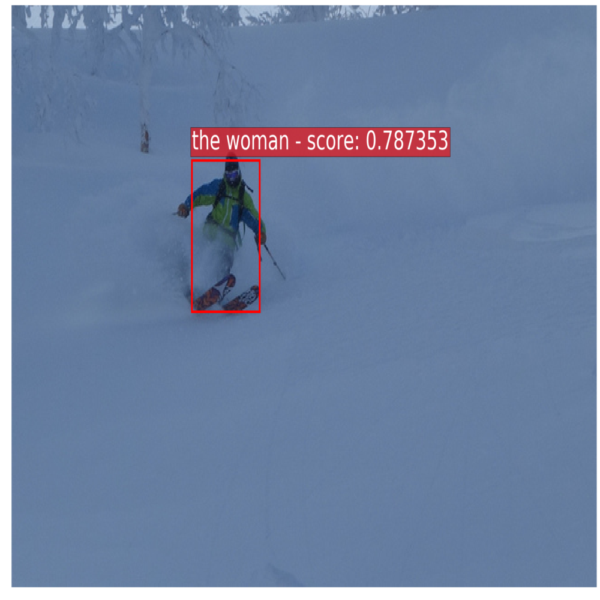}

{\footnotesize what is [the woman] riding\par}
\end{minipage}
\end{minipage}
\end{minipage}

\begin{minipage}{1\textwidth}
\centering

\begin{minipage}{0.49\textwidth}
\centering
\begin{minipage}{0.47\textwidth}
\centering
\includegraphics[width=\linewidth]{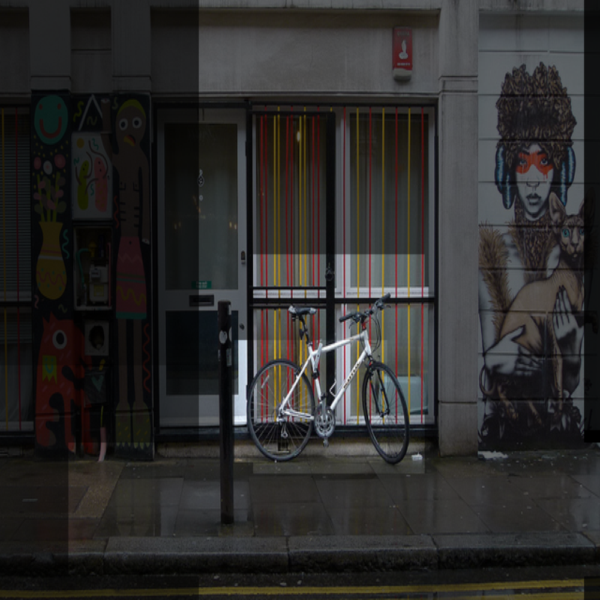}

{\footnotesize {\color[rgb]{1,0.9,0.9}is} {\color[rgb]{1,0,0}anyone} {\color[rgb]{1,0.6,0.6}riding} {\color[rgb]{1,0.3,0.3}the bike}\par}
{\footnotesize matching score: 3.573e-07\par}
\end{minipage}
\begin{minipage}{0.47\textwidth}
\centering
\includegraphics[width=\linewidth]{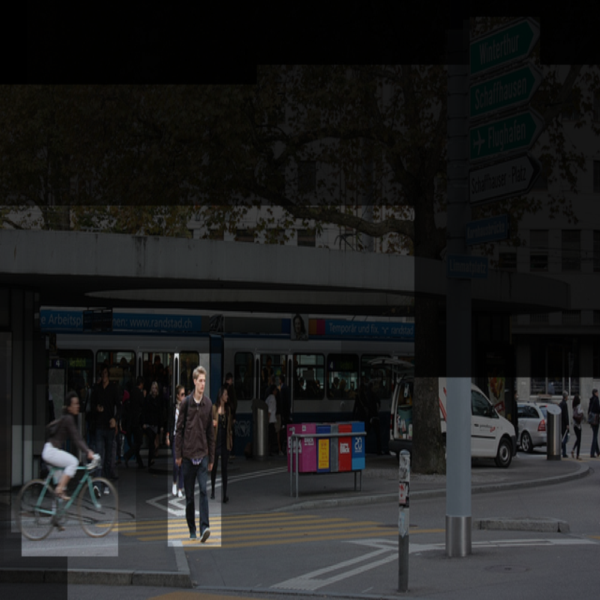}

{\footnotesize {\color[rgb]{1,0.9,0.9}is} {\color[rgb]{1,0,0}anyone} {\color[rgb]{1,0.6,0.6}riding} {\color[rgb]{1,0,0}the bike}\par}
{\footnotesize matching score: 0.387\par}
\end{minipage}
\end{minipage}
\begin{minipage}{0.49\textwidth}
\centering
\begin{minipage}{0.47\textwidth}
\centering
\includegraphics[width=\linewidth]{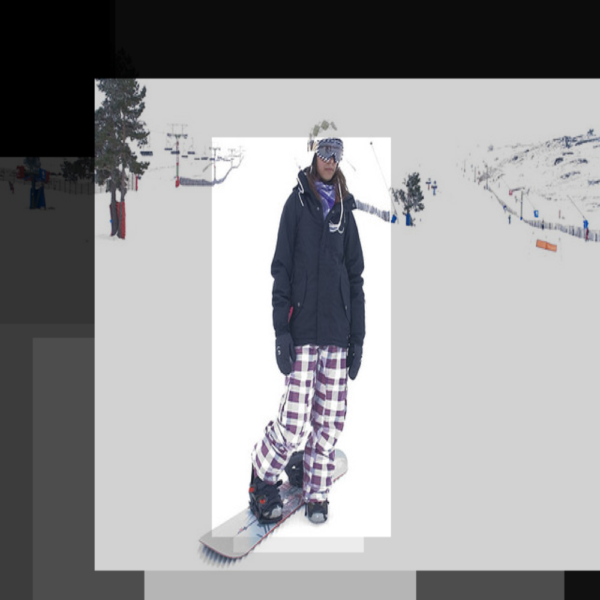}

{\footnotesize {\color[rgb]{1,0.8,0.8}what} {\color[rgb]{1,0.3,0.3}is} {\color[rgb]{1,0.9,0.9}the} {\color[rgb]{1,0.3,0.3}woman} {\color[rgb]{1,0.4,0.4}riding}\par}
{\footnotesize matching score: 0.083\par}
\end{minipage}
\begin{minipage}{0.47\textwidth}
\centering
\includegraphics[width=\linewidth]{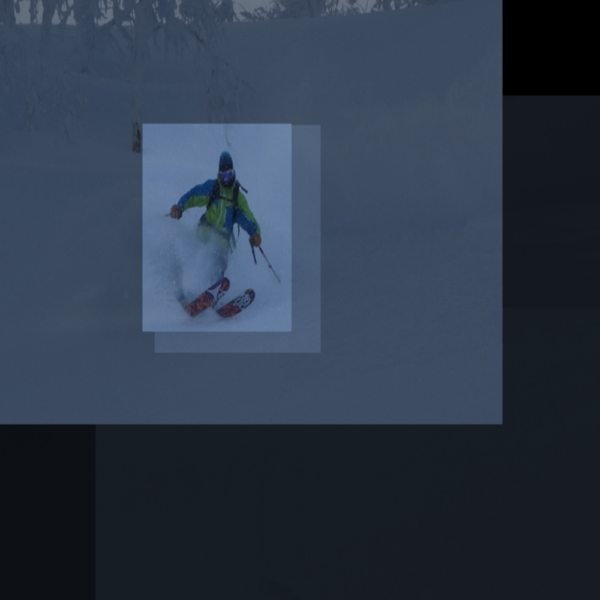}

{\footnotesize {\color[rgb]{1,0.8,0.8}what} {\color[rgb]{1,0.3,0.3}is} {\color[rgb]{1,0.9,0.9}the} {\color[rgb]{1,0,0}woman} {\color[rgb]{1,0.4,0.4}riding}\par}
{\footnotesize matching score: 0.008\par}
\end{minipage}
\end{minipage}
\end{minipage}

\begin{minipage}{1\textwidth}
\centering

\begin{minipage}{0.49\textwidth}
\centering
\begin{minipage}{0.47\textwidth}
\centering
\includegraphics[width=\linewidth]{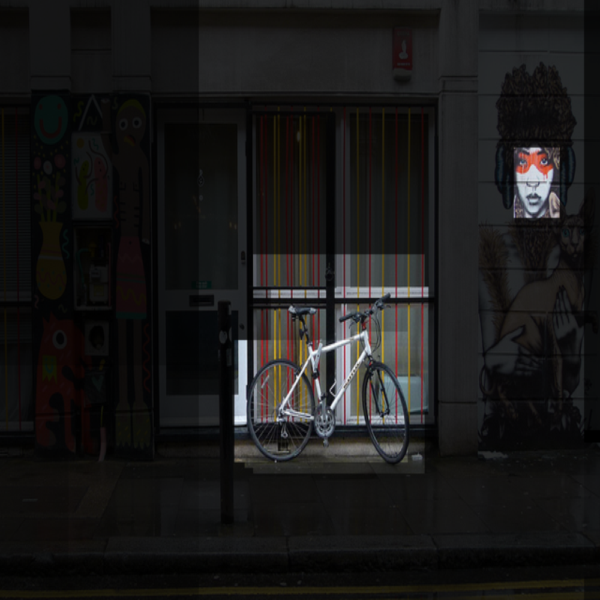}

{\footnotesize {\color[rgb]{1,0,0}is} {\color[rgb]{1,0.3,0.3}anyone} {\color[rgb]{1,0.6,0.6}riding} {\color[rgb]{1,0.6,0.6}the bike}\par}

{\footnotesize Pred: no, Ans: no\par}
\end{minipage}
\begin{minipage}{0.47\textwidth}
\centering
\includegraphics[width=\linewidth]{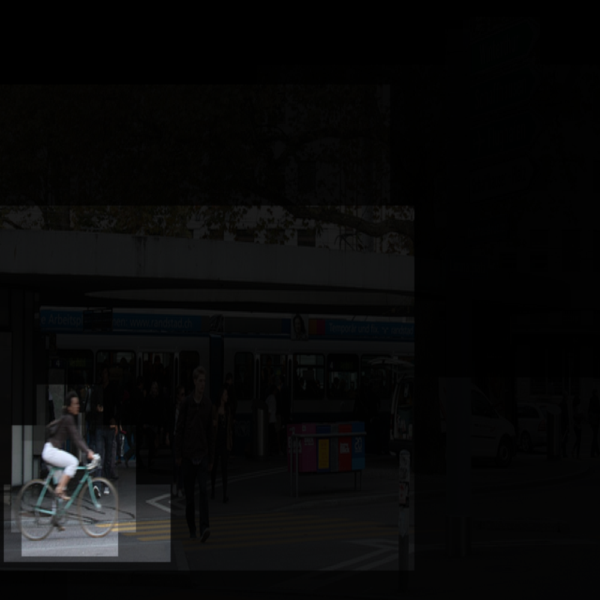}

{\footnotesize {\color[rgb]{1,0,0}is} {\color[rgb]{1,0,0}anyone} {\color[rgb]{1,0.6,0.6}riding} {\color[rgb]{1,0.6,0.6}the bike}\par}

{\footnotesize Pred: yes, Ans: yes\par}
\end{minipage}
\end{minipage}
\begin{minipage}{0.49\textwidth}
\centering
\begin{minipage}{0.47\textwidth}
\centering
\includegraphics[width=\linewidth]{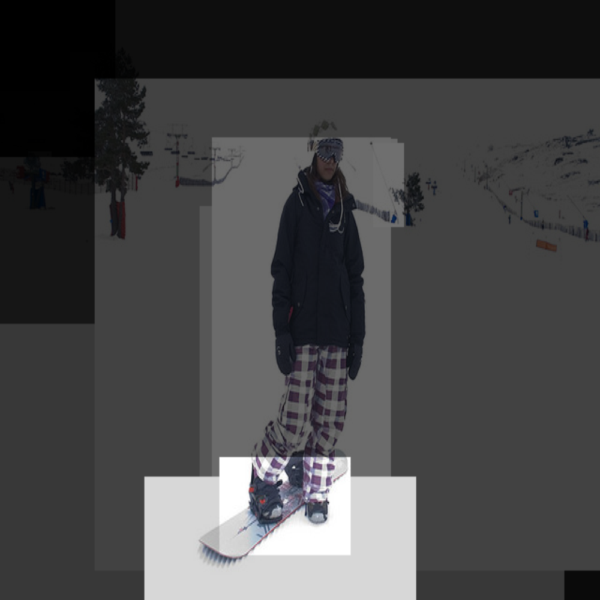}

{\footnotesize {\color[rgb]{1,0,0}what} {\color[rgb]{1,0.6,0.6}is} {\color[rgb]{1,0.9,0.9}the} {\color[rgb]{1,0.6,0.6}woman} {\color[rgb]{1,0.6,0.6}riding}\par}

{\footnotesize Pred: snowboard, Ans: snowboard\par}
\end{minipage}
\begin{minipage}{0.47\textwidth}
\centering
\includegraphics[width=\linewidth]{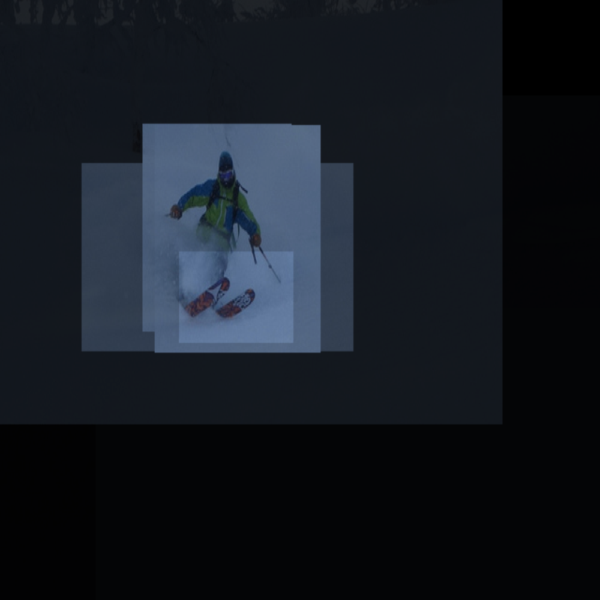}

{\footnotesize {\color[rgb]{1,0,0}what} {\color[rgb]{1,0.6,0.6}is} {\color[rgb]{1,0.6,0.6}the} {\color[rgb]{1,0.9,0.9}woman} {\color[rgb]{1,0.6,0.6}riding}\par}

{\footnotesize Pred: skis, Ans: skis\par}
\end{minipage}
\end{minipage}
\end{minipage}

\caption{Example visualizations of behaviours of our network for two complementary image-question pairs (i.e., samples with the same question but different images and answers) from VQA 2.0 dataset. The three rows (from top to bottom) show the behaviours of the VG, ICR, and VQA decoders, respectively. For VG, top-1 regions corresponding to the entities (i.e., NP chunks) in the questions are shown. For ICR and VQA, the attention maps generated in their decoders are shown; the brightness of image pixels and the redness of words indicate the attention weights.
}
\label{fig:visualization}
\end{figure*}

To evaluate the effectiveness of joint learning, we first trained the model  on all possible combinations out of the three tasks and evaluated their performances To be specific, for each  combination of tasks, we trained our mode on their \textit{train} split(s) and calculating its performance for each of the trained tasks on its \textit{val} split. When training on two or more tasks, we used the method explained in Sec.~\ref{sec:taskschedule}.

Table \ref{table:ablation} shows the results. It is observed that the joint learning of two tasks achieves better or comparable performances than the learning on a single task; and that the joint learning of all the three tasks yields the best performance. These confirm the effectiveness of our method for multi-task learning.

For ICR, we use two datasets, MS-COCO and Flickr30k; the former is about three times larger than the latter. We evaluated how performances vary between when using the former and when using the latter. Table \ref{table:compare} shows the results. It is observed that 
the joint learning with VQA and VG is more beneficial for the smaller dataset (Flickr30k) than the larger one (MS-COCO), e.g., from 67.16 to 72.07 vs. from 69.05 to 70.43 (ICR: image annotation). On the other hand, the improvements of the other tasks (VQA and VG) due to the joint training with ICR are smaller for Flickr30k than for MS-COCO, e.g., from 65.50 to 66.09 vs. 65.50 to 66.35 (VQA).

\begin{table}
\begin{center}
\caption{Comparison of our method and previous ones on the visual grounding task using Flickr30k Entities in the same condition.}
\label{table:res_vg} \footnotesize
\begin{tabular}{|c|ccc|}
\hline
Method & R@1 & R@5 & R@10 \\
\hline
\hline
Top-Down Attention \cite{Zhang_2016_ECCV} & 28.50 & 52.70 & 61.30 \\
SCRC \cite{Hu_2016_CVPR} & 27.80 & - & 62.90 \\
Structured Matching \cite{Wang_2016_ECCV} & 42.08 & - & - \\
DSPE \cite{Wang_2016_CVPR} & 43.89 & 64.46 & 68.66 \\
GroundeR \cite{Rohrbach_2016_ECCV} & 48.38 & - & - \\
MCB \cite{Fukui_2016_EMNLP} & 48.69 & - & - \\
RTP \cite{Plummer_2017_IJCV} & 50.89 & \bf{71.09} & \bf{75.73} \\
GOP \cite{Yeh_2017_NIPS} & 53.97 & - & - \\
Ours & \bf{57.39} & 69.37 & 71.03 \\
\hline
\end{tabular}
\end{center}
\end{table}

\subsection{Full Results on Test Sets}
\label{sec:results}

We next show the performance of our method on \emph{test} sets for the three tasks. We employ the procedure for avoiding training data contamination explained in the last paragraph of Sec.~\ref{sec:data}. We show below comparisons of our method with previous methods on each task. Note that our method alone performs joint learning of the three tasks and others are trained only on each individual task. 

\paragraph{Image Caption Retrieval} Table \ref{table:res_icr} shows the performances of previous methods and our method on Flickr30k and MS-COCO (The numbers for MS-COCO are performance on the 1,000 testing images. In the supplementary material, we report the performance on the 5,000 testing images of MS-COCO). It is seen that our method is comparable with the state-of-the-art method (S-E Model) on MS-COCO. For Flickr30k, which is three times smaller than MS-COCO, our method outperforms the best published result (S-E Model) by a large margin  (about 9.5\% in average) on all six evaluation criteria, showing the effectiveness of our method. This demonstrates that our method can leverage the joint learning with other tasks to cover insufficient amount of training data for ICR.

\paragraph{Visual Question Answering} Table \ref{table:res_vqa2.0} shows comparisons of our method to previous published results on VQA 2.0 in both test-dev and test-standard sets. It is observed in Table \ref{table:res_vqa2.0} that our method outperforms the state-of-the-art  method (DCN \cite{Nguyen_2018_CVPR}) by a noticable margin of $\sim 0.7\%$ on the two test sets. It is noted that the improvements are seen in all the question types of test-standard set (\textit{Other} with $0.5 \%$, \textit{Number} with $0.2 \%$, and \textit{Yes/No} with $0.9 \%$). Notably, its accuracy for counting questions (\textit{Number}) is on par with the Counting Module, which is designed to improve accuracy of this question type.

\paragraph{Visual Grounding} Table \ref{table:res_vg} shows comparisons of our method with previous methods on the Flickr30k Entities dataset. Although our method shows lower performance than RTP \cite{Plummer_2017_IJCV} on the R@5 and R@10 evaluation metrics, it achieves a much better result on the hardest metric R@1. It should be noted that our method uses only phrase-box correspondences provided in the training dataset, and does not use any other information, such as box size, color, segmentation, or pose-estimation, which are used in previous studies \cite{Plummer_2017_IJCV, Yeh_2017_NIPS}.

\subsection{Qualitative Evaluation}

To analyze effects of joint learning of multiple tasks, we visualize behaviours of our network. We use complementary image-question pairs contained in VQA 2.0 \cite{balanced_vqa_v2} for better analyses. Figure \ref{fig:visualization} shows two examples of such visualization, each for a complementary image-question pair. For visualization of VG, we extract NP chunks from the input question and treat them as entities. We then compute the score between each entity and all of the image regions, as described in Sec.~\ref{subsec:visual_grounding}. The first row of Fig.~\ref{fig:visualization} shows the correspondences between a few entities found in the questions and their top-1 image regions. For ICR and VQA, we visualize attention maps generated in their decoders, which are shown in the second and third rows of Fig.~\ref{fig:visualization}.

It can be seen from the first row of Fig.~\ref{fig:visualization} that the VG decoder correctly aligns each entity to its corresponding image region. From the second row of Fig.~\ref{fig:visualization} (i.e., the attention maps of the ICR decoder) we can observe that the ICR decoder is looking at the same entities as those found in the VG decoder but with wider attention in the image and sentence, implying that not only the relevant entities but their relations are captured in the ICR decoder. It is then seen from the third row of Fig.~\ref{fig:visualization} (i.e., the attention weights on image regions and question words of the VQA decoder) that it narrows down its attention on the image regions and question words that are relevant to properly answer the input questions, e.g., the bikes in the images and the phrase ``\textit{is enyone}'' in the questions; and the snowboard and the skis in the images and the phrase ``\textit{what is}''.

Other observations can be made for the results in Fig.~\ref{fig:visualization}. For instance, the ICR decoder gives a very low score (\texttt{3.573e-07}) for the pair of the first image and the question ``\textit{is anyone riding the bike}'' and a high score (\texttt{0.387}) for  the second image and the same question. Considering the word attention  focusing only on the phrase ''\textit{anyone riding the bike}'', we may think that the ICR decoder  correctly judges the (in)consistency between the contents of the images and the phrase. These agree well with their correct answers in VQA (i.e., ``\textit{No}'' and ``\textit{Yes}''), implying the interaction between ICR and VQA. Further analyses will be provided in supplementary material.

\section{Summary and Conclusion}

In this paper, we have presented a multi-task learning framework for
vision-language tasks. The key component is the proposed network
consisting of the representation encoder that learns to fuse visual and linguistic representations in a hierarchical fashion, and task-specific decoders that utilize the learned representation at their corresponding levels in the hierarchy to make prediction. We have shown the effectiveness of our approach through a series of experiments on three major tasks and their datasets. The shared hierarchical representation learned by the encoder has been shown to generalize well across the tasks.

{\small
\bibliographystyle{ieee}
\bibliography{camera_ready}

\begin{thebibliography}{10}\itemsep=-1pt

\bibitem{Uddin_2018_ICLR}
W.~U. Ahmad, K.-W. Chang, and H.~Wang.
\newblock Multi-task learning for document ranking and query suggestion.
\newblock In {\em International Conference on Learning Representations (ICLR)},
  2018.

\bibitem{Anderson_2018_CVPR}
P.~Anderson, X.~He, C.~Buehler, D.~Teney, M.~Johnson, S.~Gould, and L.~Zhang.
\newblock Bottom-up and top-down attention for image captioning and visual
  question answering.
\newblock In {\em International Conference on Computer Vision and Pattern
  Recognition (CVPR)}, 2018.

\bibitem{VQA}
S.~Antol, A.~Agrawal, J.~Lu, M.~Mitchell, D.~Batra, C.~L. Zitnick, and
  D.~Parikh.
\newblock {VQA}: {V}isual {Q}uestion {A}nswering.
\newblock In {\em International Conference on Computer Vision (ICCV)}, 2015.

\bibitem{Bai_2018_ECCV}
Y.~Bai, J.~Fu, T.~Zhao, and T.~Mei.
\newblock Deep attention neural tensor network for visual question answering.
\newblock In {\em European Conference on Computer Vision (ECCV)}, 2018.

\bibitem{Caruana1997}
R.~Caruana.
\newblock Multitask learning.
\newblock {\em Machine Learning}, 1997.

\bibitem{Chen_2017_ICCV}
Z.~Chen, Z.~Yanpeng, H.~Shuaiyi, T.~Kewei, and M.~Yi.
\newblock Structured attentions for visual question answering.
\newblock In {\em International Conference on Computer Vision (ICCV)}, 2017.

\bibitem{visdial}
A.~Das, S.~Kottur, K.~Gupta, A.~Singh, D.~Yadav, J.~M. Moura, D.~Parikh, and
  D.~Batra.
\newblock {V}isual {D}ialog.
\newblock In {\em International Conference on Computer Vision and Pattern
  Recognition (CVPR)}, 2017.

\bibitem{Dong_2015_IJCNLP}
D.~Dong, H.~Wu, W.~He, D.~Yu, and H.~Wang.
\newblock Multi-task learning for multiple language translation.
\newblock In {\em International Joint Conference on Natural Language Processing
  (IJCNLP)}, 2015.

\bibitem{Aviv_2017_CVPR}
A.~Eisenschtat and L.~Wolf.
\newblock Capturing deep correlations with 2-way nets.
\newblock In {\em International Conference on Computer Vision and Pattern
  Recognition (CVPR)}, 2017.

\bibitem{Faghri_2017_arxiv}
F.~Faghri, D.~J. Fleet, R.~Kiros, and S.~Fidler.
\newblock {VSE++:} improved visual-semantic embeddings.
\newblock {\em arXiv preprint arXiv:1707.05612}, 2017.

\bibitem{Fukui_2016_EMNLP}
A.~Fukui, D.~H. Park, D.~Yang, A.~Rohrbach, T.~Darrell, and M.~Rohrbach.
\newblock Multimodal compact bilinear pooling for visual question answering and
  visual grounding.
\newblock In {\em Empirical Methods in Natural Language Processing (EMNLP)},
  2016.

\bibitem{Glorot_2010}
X.~Glorot and Y.~Bengio.
\newblock Understanding the difficulty of training deep feedforward neural
  networks.
\newblock In {\em International Conference on Artificial Intelligence and
  Statistics}, 2010.

\bibitem{balanced_vqa_v2}
Y.~Goyal, T.~Khot, D.~Summers{-}Stay, D.~Batra, and D.~Parikh.
\newblock Making the {V} in {VQA} matter: Elevating the role of image
  understanding in {V}isual {Q}uestion {A}nswering.
\newblock In {\em International Conference on Computer Vision and Pattern
  Recognition (CVPR)}, 2017.

\bibitem{Hu_2016_CVPR}
R.~Hu, H.~Xu, M.~Rohrbach, J.~Feng, K.~Saenko, and T.~Darrell.
\newblock Natural language object retrieval.
\newblock In {\em International Conference on Computer Vision and Pattern
  Recognition (CVPR)}, 2016.

\bibitem{Huang_2017_CVPR}
Y.~Huang, W.~Wang, and L.~Wang.
\newblock Instance-aware image and sentence matching with selective multimodal
  {LSTM}.
\newblock In {\em International Conference on Computer Vision and Pattern
  Recognition (CVPR)}, 2016.

\bibitem{Huang_2018_CVPR}
Y.~Huang, Q.~Wu, C.~Song, and L.~Wang.
\newblock Learning semantic concepts and order for image and sentence matching.
\newblock In {\em International Conference on Computer Vision and Pattern
  Recognition (CVPR)}, 2018.

\bibitem{Karpathy_2015_CVPR}
A.~Karpathy and L.~Fei-Fei.
\newblock Deep visual-semantic alignments for generating image descriptions.
\newblock In {\em International Conference on Computer Vision and Pattern
  Recognition (CVPR)}, 2015.

\bibitem{Kendall_2018_CVPR}
A.~Kendall, Y.~Gal, and R.~Cipolla.
\newblock Multi-task learning using uncertainty to weigh losses for scene
  geometry and semantics.
\newblock In {\em International Conference on Computer Vision and Pattern
  Recognition (CVPR)}, 2018.

\bibitem{Lev_2016_ECCV}
G.~Lev, G.~Sadeh, B.~Klein, and L.~Wolf.
\newblock {RNN} fisher vectors for action recognition and image annotation.
\newblock In {\em European Conference on Computer Vision (ECCV)}, 2016.

\bibitem{Li_2018_CVPR}
Y.~Li, N.~Duan, B.~Zhou, X.~Chu, W.~Ouyang, X.~Wang, and M.~Zhou.
\newblock Visual question generation as dual task of visual question answering.
\newblock In {\em International Conference on Computer Vision and Pattern
  Recognition (CVPR)}, 2018.

\bibitem{Li_2017_CVPR}
Y.~Li, W.~Ouyang, B.~Zhou, K.~Wang, and X.~Wang.
\newblock Scene graph generation from objects, phrases and region captions.
\newblock In {\em International Conference on Computer Vision and Pattern
  Recognition (CVPR)}, 2017.

\bibitem{mscoco}
T.-Y. Lin, M.~Maire, S.~Belongie, J.~Hays, P.~Perona, D.~Ramanan,
  P.~Doll{\'a}r, and C.~L. Zitnick.
\newblock Microsoft coco: Common objects in context.
\newblock In {\em European Conference on Computer Vision (ECCV)}, 2014.

\bibitem{Lin_2016_ECCV}
X.~Lin and D.~Parikh.
\newblock Leveraging visual question answering for image-caption ranking.
\newblock In {\em European Conference on Computer Vision (ECCV)}, 2016.

\bibitem{Liu_2017_ICCV}
Y.~Liu, Y.~Guo, E.~M. Bakker, and M.~S. Lew.
\newblock Learning a recurrent residual fusion network for multimodal matching.
\newblock In {\em International Conference on Computer Vision (ICCV)}, 2017.

\bibitem{Lu_2018_CVPR}
J.~Lu, J.~Yang, D.~Batra, and D.~Parikh.
\newblock Neural baby talk.
\newblock In {\em International Conference on Computer Vision and Pattern
  Recognition (CVPR)}, 2018.

\bibitem{Luong_2016_ICLR}
M.~Luong, Q.~V. Le, I.~Sutskever, O.~Vinyals, and L.~Kaiser.
\newblock Multi-task sequence to sequence learning.
\newblock In {\em International Conference on Learning Representations (ICLR)},
  2016.

\bibitem{Ma_2018_CVPR}
C.~Ma, C.~Shen, A.~R. Dick, and A.~van~den Hengel.
\newblock Visual question answering with memory-augmented networks.
\newblock In {\em International Conference on Computer Vision and Pattern
  Recognition (CVPR)}, 2018.

\bibitem{Hyeonseob_2017_CVPR}
H.~Nam, J.~Ha, and J.~Kim.
\newblock Dual attention networks for multimodal reasoning and matching.
\newblock In {\em International Conference on Computer Vision and Pattern
  Recognition (CVPR)}, 2017.

\bibitem{Nguyen_2018_CVPR}
D.-K. Nguyen and T.~Okatani.
\newblock Improved fusion of visual and language representations by dense
  symmetric co-attention for visual question answering.
\newblock In {\em International Conference on Computer Vision and Pattern
  Recognition (CVPR)}, 2018.

\bibitem{Pennington_2014_EMNLP}
J.~Pennington, R.~Socher, and C.~D. Manning.
\newblock Glove: Global vectors for word representation.
\newblock In {\em Empirical Methods in Natural Language Processing (EMNLP)},
  2014.

\bibitem{Plummer_2017_IJCV}
B.~A. Plummer, L.~Wang, C.~M. Cervantes, J.~C. Caicedo, J.~Hockenmaier, and
  S.~Lazebnik.
\newblock Flickr30k entities: Collecting region-to-phrase correspondences for
  richer image-to-sentence models.
\newblock {\em International Journal of Computer Vision (IJCV)}, 2017.

\bibitem{Ray_2018_ECCV}
J.~Ray, H.~Wang, D.~Tran, Y.~Wang, M.~Feiszli, L.~Torresani, and M.~Paluri.
\newblock Scenes-objects-actions: A multi-task, multi-label video dataset.
\newblock In {\em European Conference on Computer Vision (ECCV)}, 2018.

\bibitem{Rohrbach_2016_ECCV}
A.~Rohrbach, M.~Rohrbach, R.~Hu, T.~Darrell, and B.~Schiele.
\newblock Grounding of textual phrases in images by reconstruction.
\newblock In {\em European Conference on Computer Vision (ECCV)}, 2016.

\bibitem{Su_2018_CVPR}
Z.~Su, C.~Zhu, Y.~Dong, D.~Cai, Y.~Chen, and J.~Li.
\newblock Learning visual knowledge memory networks for visual question
  answering.
\newblock In {\em International Conference on Computer Vision and Pattern
  Recognition (CVPR)}, 2018.

\bibitem{Subramanian_2018_ICLR}
S.~Subramanian, A.~Trischler, Y.~Bengio, and C.~J. Pal.
\newblock Learning general purpose distributed sentence representations via
  large scale multi-task learning.
\newblock In {\em International Conference on Learning Representations (ICLR)},
  2018.

\bibitem{Teney_2018_CVPR}
D.~Teney, P.~Anderson, X.~He, and A.~van~den Hengel.
\newblock Tips and tricks for visual question answering: Learnings from the
  2017 challenge.
\newblock In {\em International Conference on Computer Vision and Pattern
  Recognition (CVPR)}, 2018.

\bibitem{Ivan_2016_ICLR}
I.~Vendrov, R.~Kiros, S.~Fidler, and R.~Urtasun.
\newblock Order-embeddings of images and language.
\newblock In {\em International Conference on Learning Representations (ICLR)},
  2016.

\bibitem{Wang_2016_CVPR}
L.~Wang, Y.~Li, and S.~Lazebnik.
\newblock Learning deep structure-preserving image-text embeddings.
\newblock In {\em International Conference on Computer Vision and Pattern
  Recognition (CVPR)}, 2016.

\bibitem{Wang_2016_ECCV}
M.~Wang, M.~Azab, N.~Kojima, R.~Mihalcea, and J.~Deng.
\newblock Structured matching for phrase localization.
\newblock In {\em European Conference on Computer Vision (ECCV)}, 2016.

\bibitem{Yeh_2017_NIPS}
R.~Yeh, J.~Xiong, W.-M. Hwu, M.~Do, and A.~Schwing.
\newblock Interpretable and globally optimal prediction for textual grounding
  using image concepts.
\newblock In {\em Advances in Neural Information Processing Systems (NIPS)},
  2017.

\bibitem{Young_2014_TACL}
P.~Young, A.~Lai, M.~Hodosh, and J.~Hockenmaier.
\newblock From image descriptions to visual denotations: New similarity metrics
  for semantic inference over event descriptions.
\newblock In {\em Transactions of the Association for Computational Linguistics
  (TACL)}, 2014.

\bibitem{Zhang_2016_ECCV}
J.~Zhang, S.~A. Bargal, Z.~Lin, J.~Brandt, X.~Shen, and S.~Sclaroff.
\newblock Top-down neural attention by excitation backprop.
\newblock In {\em European Conference on Computer Vision (ECCV)}, 2016.

\bibitem{Zhang_2018_ICLR}
Y.~Zhang, J.~Hare, and A.~Prügel-Bennett.
\newblock Learning to count objects in natural images for visual question
  answering.
\newblock In {\em International Conference on Learning Representations (ICLR)},
  2018.

\end{thebibliography}
}

\end{document}